\documentclass[runningheads]{llncs}
\usepackage{graphicx}
\usepackage[none]{hyphenat}
\begin{document}
\sloppy

\title{Real-time Embedded Person Detection and Tracking for Shopping Behaviour Analysis}

\titlerunning{Real-time Embedded Person Detection and Tracking}

\author{R. Schrijvers\inst{1,2} \and S. Puttemans\inst{1} 
\and T. Callemein\inst{1} \and T. Goedem\'{e}\inst{1}}

\authorrunning{R. Schrijvers; S. Puttemans; et al.}

\institute{EAVISE, KU Leuven, Jan De Nayerlaan 5, 2860 Sint-Katelijne-Waver, Belgium \\ \and
PXL Smart-ICT, Hogeschool PXL, Elfde Liniestraat 24, 35000 Hasselt, Belgium\\
\email{\{steven.puttemans, toon.goedeme, timothy.callemein\}@kuleuven.be}\\
\email{robin.schrijvers@pxl.be}\\} 

\maketitle

\begin{abstract}
Shopping behaviour analysis through counting and tracking of people in shop-like environments offers valuable information for store operators and provides key insights in the stores layout (e.g. frequently visited spots). Instead of using extra staff for this, automated on-premise solutions are preferred. These automated systems should be cost-effective, preferably on lightweight embedded hardware, work in very challenging situations (e.g. handling occlusions) and preferably work real-time. We solve this challenge by implementing a real-time TensorRT optimized YOLOv3-based pedestrian detector, on a Jetson TX2 hardware platform. By combining the detector with a sparse optical flow tracker we assign a unique ID to each customer and tackle the problem of loosing partially occluded customers. Our detector-tracker based solution achieves an average precision of 81.59\% at a processing speed of 10 FPS. Besides valuable statistics, heat maps of frequently visited spots are extracted and used as an overlay on the video stream.

\end{abstract}

\section{Introduction} \label{introduction}
Mapping the flow and deriving statistics (e.g. the amount of visitors or the time spent in as store) of people visiting shop-like environments, holds high value for store operators. To this day, people counting in retail environments is often being accomplished by using cross-line detection systems \cite{linedetection} or algorithms counting people through virtual gates \cite{virtualgate}.

\begin{figure}[!t]
 \centering
  \begin{minipage}[b]{0.49\textwidth}
    \includegraphics[width=\textwidth]{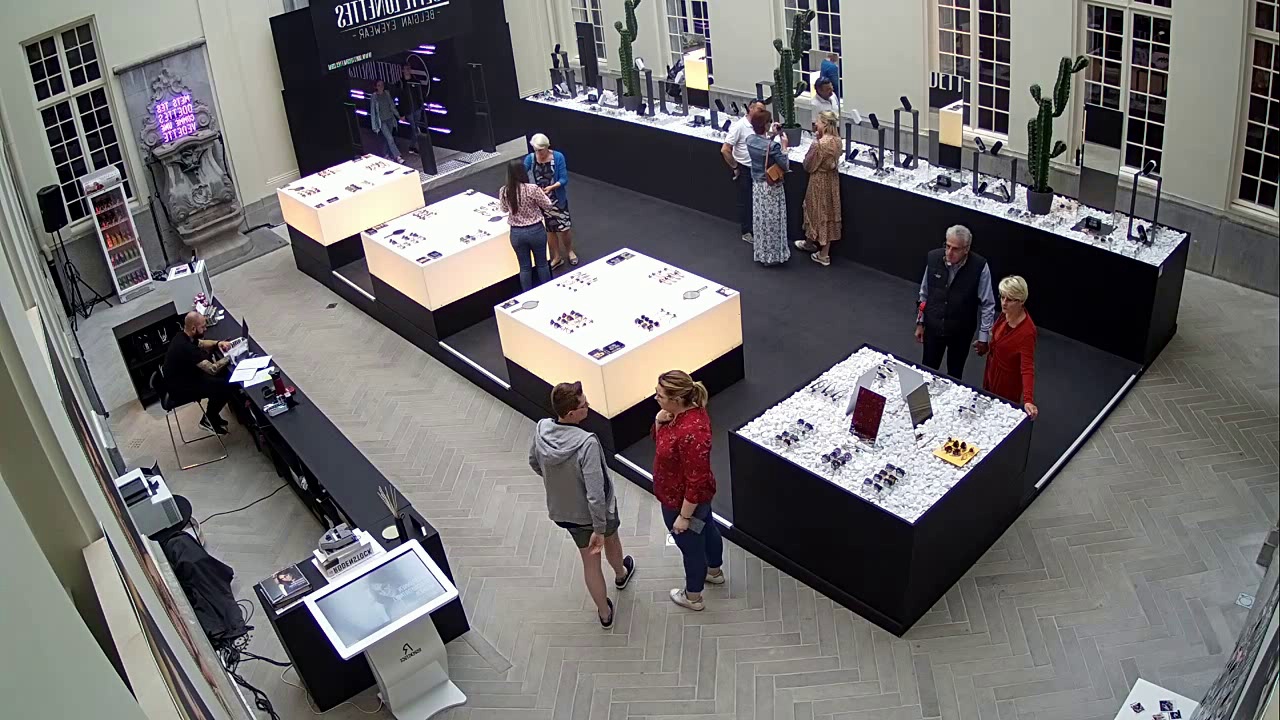}
  \end{minipage}
  \begin{minipage}[b]{0.49\textwidth}
    \includegraphics[width=\textwidth]{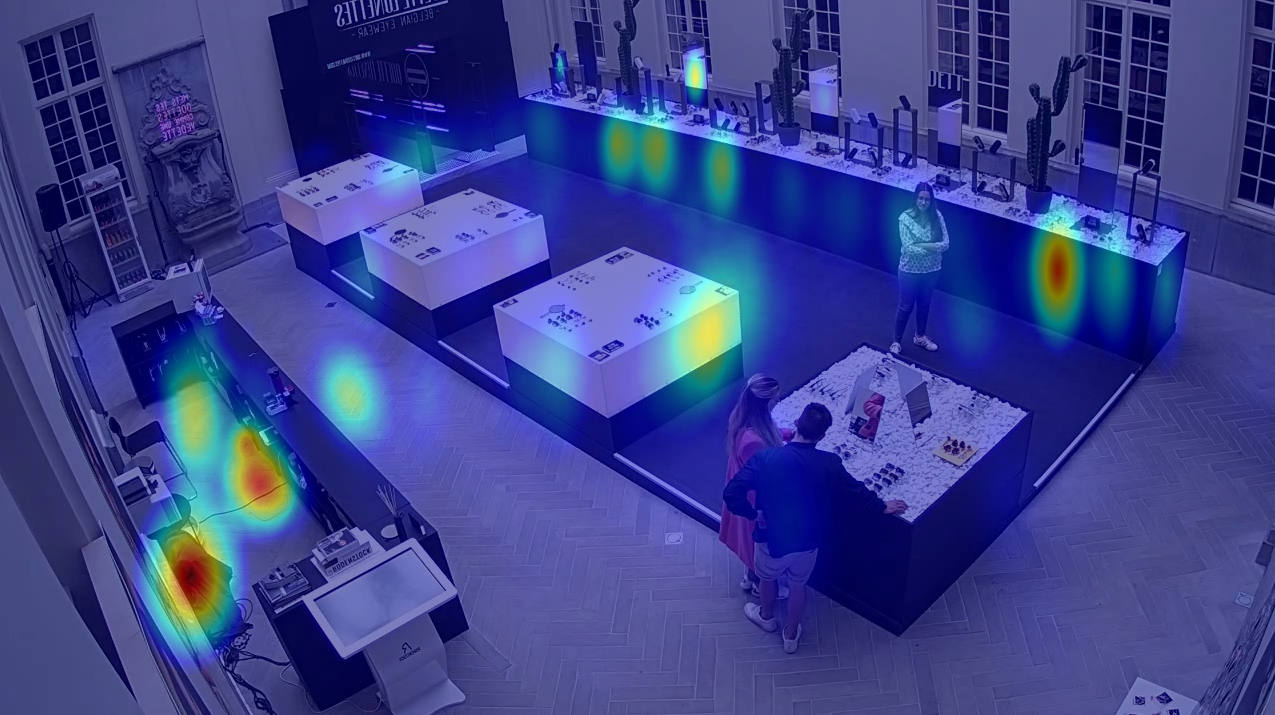}
  \end{minipage}
  \caption{Examples of (left) the viewpoint from our camera setup in the store and (right) the generated heat map for the store owner using the combined detector-tracker unit.}
  \label{fig:heatmaps}
\end{figure}

To accurately count visitors, one can place a computer in the network of the store with access to already available security cameras, deploying software that detects and tracks people in the store automatically and stores the results on a central storage system (e.g. an in-store server or the cloud). In order to run these software solutions, one needs expensive and bulky dedicated computing hardware, frequently covered by a desktop GPU (e.g. NVIDIA RTX 2080). On the other hand, the recent availability of lightweight and affordable embedded GPU solutions, like the NVIDIA Jetson TX2, can be a valid alternative. This is the main motivator to build an embedded and cost-effective people detection and tracking solution. An example of the camera viewpoint from the designed setup can be seen in the left part of Figure \ref{fig:heatmaps}. 

The remainder of this paper is organized as follows. Section \ref{relatedwork} discusses related work on person detection and tracking, along with available embedded hardware solutions. Section \ref{methodology} provides details about the proposed implementation, while experiments and results are discussed in section \ref{results}. Finally, section \ref{conclusion} summarizes this work and discusses useful future research directions.

\section{Related Work} \label{relatedwork}

The related work section is subdivided in three subsections, each focusing on a specific subtopic within this manuscript. Subsection \ref{persondetection} starts by discussing literature on person detection, where subsection \ref{embeddedhardware} continues on specific optimizations in this technology for embedded hardware. Finally subsection \ref{persontracking} focuses on the person tracking part.

\subsection{Person Detection} \label{persondetection}

Robust person detection solutions have been heavily studied in literature. Originally person detection made use of handcrafted features, combined with machine learning to generate an abstract representation of the person \cite{randomforests,HOG,violajones,dpm,acf,icf}. While these approaches showed promising results, their top-accuracy and flexibility make that these algorithms are not suited for the very challenging conditions in which they will be deployed in this application. Dynamic backgrounds, illumination changes and different store lay-outs are only some of the reasons of a high rate of false positive detections. While re-training the detectors for every store layout would theoretically solve the issue, this isn't a cost-effective solution for companies. 

Convolutional neural networks for detecting persons in images offer potentially more robust solutions \cite{resnet,ssd,yolo,yolov2,rcnn}. By automatically selecting the most discriminate feature set based on a very large set of training data, they quickly pushed the traditional approaches into the background. In literature these deep learning based approaches are subdivided in two categories. Single-shot approaches \cite{ssd,yolo,yolov2,yolov3} solve the detection task by classifying and proposing bounding boxes in a single feed-forward step through the network. Multi-stage approaches \cite{rcnn,maskrcnn,fastrcnn,fasterrcnn} include separate networks that first generate region proposals before classifying the objects inside the proposals. A more recent and promising multi-stage approach is Trident-Net \cite{tridentnet}, where scale-specific feature maps are built into the network. Single-stage approaches tend to have a more compact and faster architecture making them preferred in lightweight embedded hardware solutions.

\subsection{Optimizing for embedded hardware} \label{embeddedhardware}

Solving the task of person detection on embedded platforms has been an active research topic in recent years. A common approach is to enable deep learning on embedded platforms by studying more compact models such as Tiny-YOLOv2 \cite{yolov2}. While these compacter models are able to run at decent speeds on embedded hardware, they tend to lose some percentage in accuracy compared to their full counterparts (e.g. YOLOv2). Besides going compact, several approaches \cite{fastyolo,shufflenet,shufflenetv2} optimize the architecture further by looking at the indirect computation complexity and address efficient memory access and platform characteristics. More recent embedded object detection algorithms introduce optimized filter solutions, like depth-wise separable convolutions \cite{mobilenet} and inverted residuals \cite{mobilenetv2}, to further optimize the performance of these embedded solutions. 

With the uprising of FPGA chips for deep learning, several architectures like \cite{squeezenet} try to reduce the number of parameters of the models even further, having a 50x reduction in parameters compared to classic models like \cite{alexnet}. These specifically designed hardware chips also allow for fixed-point 16-bit optimizations through OpenCL \cite{FPGAyolov2}. While being very promising, these FPGA systems still lack flexibility. They are in most cases designed for a very specific case, and are thus no cost-effective solution for our problem.

While many of these embedded object detectors are explicitly shaped for running on embedded hardware, detection accuracies are still lower compared to their traditional full blown CNN counterparts (eg. YOLOv3). Taking into account that our solution should perform person detection at a minimal processing speed of 10 FPS and a minimal accuracy of 80\%, we opt for integrating an optimized embedded implementation of the YOLOv3 object detector \cite{jetnet} in our pipeline. The architecture smartly combines several optimizations based on mobile convolutions in PyTorch and TensorRT compilations.

The introduction of the NVIDIA Jetson embedded GPU enabled balancing of local-processing, low power consumption and throughput in an efficient way. \cite{jetsonsurvey} discusses several implementations of deep learning models on the Jetson platform while considering a range of applications, e.g. autonomous driving or traffic surveillance. The work focuses on obtaining low latency, to make detectors useful for providing valuable real-time feedback. The advantages in both FPGA and embedded GPU systems are the fact that they are substantially more space-efficient and less power-consuming than desktop GPUs.

\subsection{Person Tracking} \label{persontracking}

Tracking objects in videos has been studied across many fields. Whenever quasi-static environments occur, motion detection through robust background subtraction is used to identify moving objects in images \cite{trackingbyoptflow3,robustmultipersontracking,realtimecounting,literaturestudypeoplecounting}. More challenging cases involve dynamic environments, e.g. tracking objects in autonomous driving applications or drones. \cite{dronetrackingGPS} solves the tracking task using detection results of a lightweight object detection algorithm combined with euclidean distance equations, GPS-locations and data from an inertial measurement unit. Because research is moving more and more towards highly accurate CNN-based object detectors, a new sort of tracking is introduced, called tracking-by-detection. By calculating an intersection over union (IoU) between the detections of two consecutive frames and applying a threshold,
one can decide if we are dealing with the same object.

Where tracking-by-detection works well in many situations, there are also some downsides with these approaches. When people are missed in several subsequent frames by the initiating detector, the person gets lost during tracking. However, in our application of real-time costumer analysis, we want to keep a unique ID for each detected person, and thus need to fill in this gap automatically. \cite{supportvectortracking} solves this issue by integrating a SVM-classifier with an optical-flow-based tracker. With the uprising of CNN-based detectors, CNN-based object trackers have also been proposed. Limb-based detection with tracking-by-detection is proposed in \cite{trackingbydetection}, while \cite{trackingbydetectionrcnn} explains an approach using IoU information from multiple object detectors. In addition, \cite{trackingbydetectionexpansion} combines location information and similarity measures to perform tracking-by-detection.

Feature-based tracking algorithms are a valid alternative. Sparse optical flow calculates the flow of objects based on a sparse feature set \cite{lukaskanade,shitomasi} and have been successfully deployed for person tracking \cite{trackingbyoptflow1,trackingbyoptflow2,trackingbyoptflow3}. Dense optical flow \cite{gunnerfarneback} calculates the flow for every point in the detection and is thus more computationally complex. \cite{objecttrackingoversampling} propose an object tracker based on weakly aligned multiple instance local features and an appearance learning algorithm. 

A final range of trackers use online learning, where a tracker learns the representation of the object on the fly through a classifier when initialized with a bounding box \cite{boosting,MIL}. \cite{KCF} learns more efficiently from these overlapping patches by using kernelized correlation filters, while \cite{medianflow} enables tracking failure detection by tracking objects both forward and backward in time, measuring and qualifying the discrepancies between the two trajectories. 

\cite{motchallenge} gives a clear general overview of state-of-the-art trackers and their accuracy on a public dataset. For this work, we integrated three object trackers: sparse optical flow, kernelized correlation filters and median flow.

This paper proposes a complete off-the-shelf solution, taking into account the fact that we work on a compact embedded system with limited power consumption, achieve real-time performance (e.g. minimally processing at 10 FPS) and obtain acceptable accuracies (e.g. over 75\%) in a single setup. While many of these sub-tasks have been discussed already in literature, to our best knowledge, there isn't a single publication that proposes such end-to-end solution for in-store customer behaviour analysis. On top of that we are combining all parts of the pipeline in a batch system that consumes the resources (both CPU and GPU) of the host system as efficient as possible. This is the biggest novelty this work introduces and will be discussed in further detail in subsection \ref{batchprocessing}.

\begin{figure}[!t]
\centering
  \includegraphics[width=0.7\textwidth]{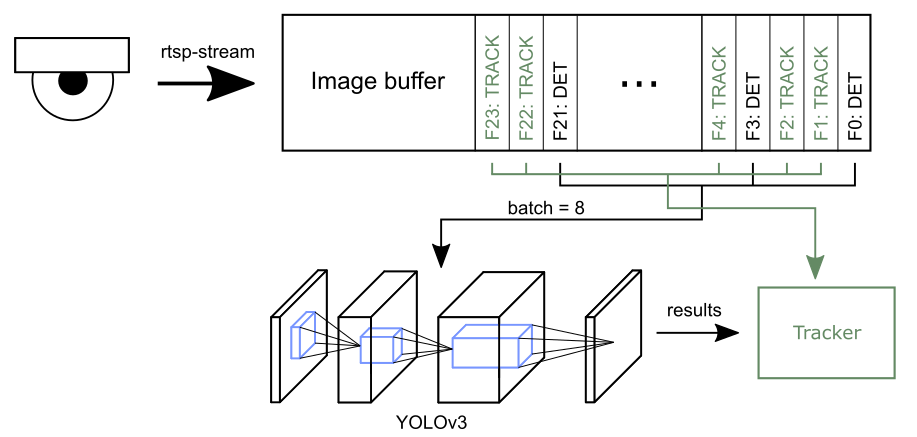}
  \caption{Pipeline overview with the proposed batch processing approach, passing frames from the image buffer to the detection or tracking unit, based on the batch iterator.}
  \label{fig:batchprocessing}

\end{figure}

\section{Methodology} \label{methodology}

The goal of this paper is to map the flow of costumers in shop-like environments, with a special focus on detecting and tracking individual costumers using a unique ID. Preferably the system should run minimally at a processing speed of 10 frames per second on top of a compact, easy-to-use, embedded platform (e.g. Jetson TX2). Therefore, we propose a multi-threaded YOLOv3 TensorRT optimized person detector combined with a fast lightweight object tracker. 

Figure \ref{fig:batchprocessing} gives an overview on how our pipeline is implemented on the embedded Jetson TX2 platform. All sub-parts will be discussed in detail in the following subsections. 

\subsection{Person detection and batch processing} \label{batchprocessing}

In applications that require real-time performance, one can either choose to maximize throughput or minimize latency. Throughput focuses on the number of images that are processed in a given time slot, while latency focuses on the time needed to process a single image. Given the goal of analyzing the flow of costumers in a given security camera stream, and in the future even in multiple camera streams, throughput is in our case more important than latency. We maximize throughput by performing a batch processing approach that takes advantage parallel GPU processing. As seen in Figure \ref{fig:batchprocessing}, an image buffer is used to collect all incoming images from the camera stream, at a 1280x720 resolution, which are downsampled to a 512$\times$288 input resolution before moving to the detector and the tracker unit. We could sent these images directly to our optimized YOLOv3 implementation \cite{jetnet}, but since it is only capable of processing data at a maximum speed of 8FPS at this resolution, we skip several frames when collecting a batch that needs to be processed by the YOLOv3 detector. 

Take for example a image buffer of 24 images. From those 24 images, only 8 images are selected as a batch that gets passed to the detector unit. The remaining 16 images are sent to the tracker unit, which waits for the processed batch of the detector, so that it can track each detected person throughout the 24 frames series, based on the detections in those 8 reference frames. In order to improve processing speed, both detector and tracker unit are implemented in separate threads, to reduce the processing time delays again as much as possible.

\subsection{Person tracking} \label{tracking}

Given our hardware setup of a Jetson TX2 embedded platform, we need to carefully consider our hardware resources. Since we're using the on-board GPU to efficiently processes batches of 8 images with the object detector unit, the decision was made to implement a CPU-based tracker to divide the workload as good as possible between the different processing units. We augment the detector unit with a lightweight CPU-based Lukas Kanade Sparse Optical Flow tracker (based on the OpenCV4.1 implementation \cite{opencvoptflow}). This allows us to run the tracker on the CPU, while simultaneously running the GPU-focused YOLOv3 object detector. To achieve the minimally required frame rate of 10 FPS, we downscale the original input frames to a resolution of 512$\times$288 pixels.

\begin{figure}[!b]
\centering
  \includegraphics[width=\textwidth]{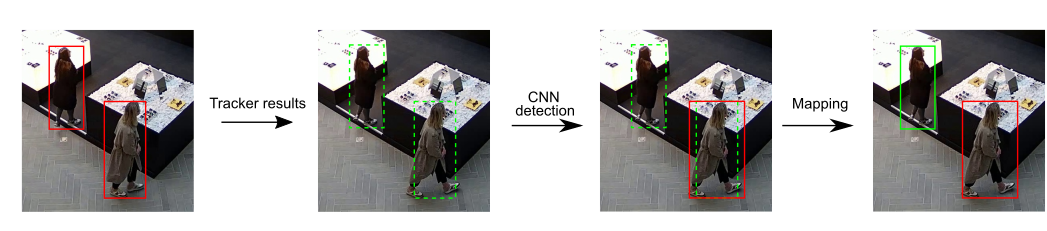}
  \caption{The detector-tracker pipeline, using tracking-by-detection with interleaved tracker proposals: (solid red) detections (interleaved green) tracker proposals (solid green) tracker proposal accepted as detection in case of a missing detection.}
  \label{fig:trackingsolution}
\end{figure}

In between detection frames, two extra frames remain which are only processed by the tracker unit, and thus so far have no knowledge of the location of the detected objects. For each detection inside a detection frame, key feature points are calculated, generating a sparse representation of that detection. Those points are passed into an optical flow mechanism that is used to generate predictions of the locations of these keypoints in the in-between tracker frames. Finally the detections in the next detection frame are used to apply a mapping between the predictions and the detections, possible proposing a slight correction of the bounding box location. Figure \ref{fig:trackingsolution} illustrates how this detector-tracker combination is working on a sample from the security camera stream. The tracker only produces predictions in the next frame for detections with a detection probability higher than 10\%. This allows us to carefully choose the working conditions of the detector and avoid as much false negative detections (persons getting missed) as possible. Detections with a lower confidence will simply be ignored by the tracker and immediately be removed from the tracker memory.

Applying this approach also introduces issues. When customers move outside the field-of-view of the camera stream, their tracking information is kept in memory by the tracker, clogging up the tracker memory and resulting in locally dangling tracks. In order to avoid these dangling tracks, after 5 consecutive detection misses, we force the tracker to remove the tracking information and forget the track altogether, which will end the tracking of that object.

Besides the sparse optical flow approach, two more CPU-based tracking implementations (Kernelized Correlation Filers and Median Flow) are tested to compare tracking robustness in shop-like environments. Both are initialized by the detections of the neural network, just like the sparse optical flow approach. For both trackers, the same input resolution of 512x288 pixels has been used.

\vspace{-2mm}
\subsection{Tracker memory} \label{memoryoftracks}

Since the selected tracker implementations only takes the information between two consecutive frames into account, we are risking losing
a person with a specific ID from tracker memory (e.g. due to exceeding the threshold of missed frames), that can in a later stage, be picked up again by the detector. In order to avoid a new ID being assigned, we introduce a method to keep and match these lost tracks in memory. In case a track is deleted from memory, it first changes its status to lost. A lost track is remembered for maximally 5 seconds, and is recovered whenever a new detection appears in a location close to the last known location of a deleted track. A match between a lost track and a newly generated track is made by meeting one of the following conditions: the new location lies within a radius of 200 pixels around the last known location, or the new location lies in the quadrant of which the gradient of the lost track points to. In case of multiple new detections in the same quadrant, the detection with the shortest (perpendicular) distance to the gradient is preferred.

\vspace{-2mm}
\subsection{Heat map extraction}

On top of providing statistics, about the exact location and path followed by customers, users of the system would also like to have some sort of visual confirmation on the activity in their stores. Therefore we propose a heat map based system that gives an overview of the store spots that are most frequently visited within a specific time slot. The heat maps are extracted by mapping the pixels in the detected and tracked bounding boxes on top of a visual layout of the store. The used object detector is known for its jittery bounding boxes, with inconsistent widths and heights. By simply incrementing pixel locations each time a detection box matches a pixel, we end up with a very jittered and visually unpleasing heat map, which we would like to avoid. We solve this by using a weighted increment of the pixels falling within a bounding box. By taking the ratio between the bounding box width and height into account and by giving the center pixel a maximum weight that degrades towards the borders of the bounding box, we end op with a increment value that is less influenced by the jitter, resulting in a visually pleasing heat map. After all frames within the given time slot have been processed, heat maps are normalized to get a meaningful color overlay, as seen in Figure \ref{fig:heatmaps}.

\section{Experiments and results} \label{results}

In this section, we evaluate the proposed embedded detector-tracker approach used for in-store customer behaviour analysis by performing four different experiments. We start by evaluating the detector unit and decide on the final deep learned architecture for the detector unit. Next, we add different trackers to the problem to cope with missing detections, optimize parameters and have a look at the influence of changing internal parameters. We then integrate our optimal detector-tracker combination into our batch processing pipeline, to gain processing speed. Finally we take a shot at generating visual heat maps, giving confirmation of frequently visited places in the stores.

\subsection{Detector and tracker evaluation}

\begin{figure}[!b]
  \centering
  \begin{minipage}[b]{0.49\textwidth}
    \includegraphics[width=\textwidth]{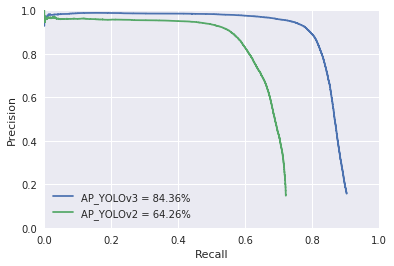}
  \end{minipage}
  \hfill
  \begin{minipage}[b]{0.49\textwidth}
    \includegraphics[width=\textwidth]{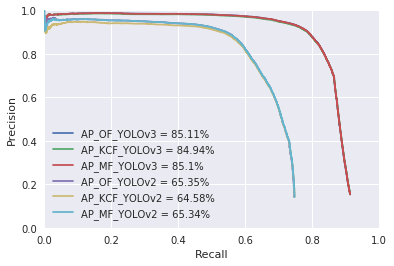}
  \end{minipage}
  \caption{Precision-recall curves of (left) the embedded and optimized YOLOv2 and YOLOv3 implementation (right) the combined detector-tracker implementations based on optical flow, median flow and kernelized correlation filters.}
  \label{fig:PRcurves}
  \vspace{-5mm}
\end{figure}

A set of 5000 in-store images were collected via the security camera streams inside the store and annotated with ground truth labels. We evaluate both the TensorRT optimized YOLOv2 and YOLOv3 implementations on this set by running them through our NVIDIA Jetson TX2 platform. The left hand side of Figure \ref{fig:PRcurves} shows the difference in obtained average precision (AP). We notice that the YOLOv3 architecture (AP=84.36\%) is performing much better than the YOLOv2 architecture (AP=64.26\%), with a tremendous rise in AP of 20\%. This can be explained by the capability of the YOLOv3 architecture to detect persons over a wider variety of scales. Giving the security cameras are mounted to give a single point overview of the store, customers walking further away from the camera are not being detected by the YOLOv2 architecture. On top of that, we also notice that the YOLOv3 architecture is better capable of detecting partially occluded persons.

However, even with the YOLOv3 architecture, we do not yet reach the optimal solution (AP=100\%). After carefully inspecting the missed detections, we notice the detect still drops persons when they are crossing each other or when they are move behind counters and thus get partially occluded. To solve these issues we run our combined detector-tracker units, for which the AP curves are visible in Figure \ref{fig:PRcurves} at the right hand side. We notice that adding the tracker unit slightly helps the detector improve in AP for all the combinations. Based on these experiments we decided to stick with the best performing combination, being YOLOv3 as detector and optical flow as tracker.

\begin{figure}[!b]
  \centering
  \begin{minipage}[b]{0.49\textwidth}
    \includegraphics[width=\textwidth]{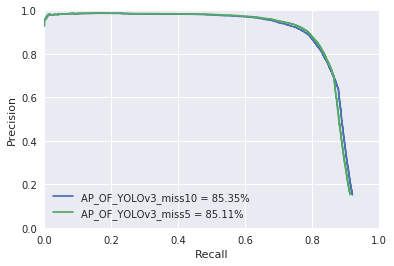}
  \end{minipage}
  \hfill
  \begin{minipage}[b]{0.49\textwidth}
    \includegraphics[width=\textwidth]{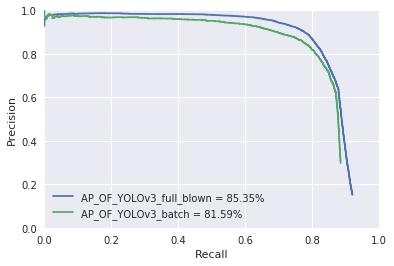}
  \end{minipage}
  \caption{Influence of (left) increasing the miss threshold for the tracker and (right) applying our batch processing pipeline to gain processing speed.}
  \label{fig:adaptations} 
\end{figure}

One could however argue that the influence of the tracker can also be increased if predicted tracks are kept in memory longer. Until now a track is rejected as soon as the detector unit is unable to find a matching detection in 5 consecutive detection frames. To show the influence of this parameter, we increased this threshold towards 10 missed detection frames and again ran the precision-recall evaluation for the optimal combination again, as seen in Figure \ref{fig:adaptations} on the left hand side. This change again provides a small boost to the AP of the optimal solution towards 85.35\%.  However, the influence of varying this parameter even further should be investigated in future research.

However, up till now our experiments were targeted at getting the highest AP possible with a detector-tracker combination. Since the original TensorRT YOLOv3 optimized implementation is only able to run the network at an average speed of 3FPS, adding the tracker reduced this further towards 2FPS. Since we want to aim for a minimal processing speed of 10 FPS, this did not suffice. As discussed in subsection \ref{batchprocessing} we applied a batch processing pipeline to overcome this low processing speed. The obtained AP is seen in the right part of Figure \ref{fig:adaptations}. While increasing our processing speed towards 9.8 FPS by simultaneously reducing the AP towards 81.59\%. In order to see if this change might influence the other detector-tracker combinations, we also evaluated those in the batch processing setup. Results of the obtained average tracking speeds can be seen in Table \ref{table:trackerspeeds}, which confirmed that we picked the correct combination before. Median filtering and kernelized correlation filters perform worse due to the linear scale-up in relation to the the amount of detected persons, while this is not true for optical flow. However, we acknowledge that this could depend on the implementation of the tracker used, and that further research on this is necessary.

\begin{table}[!t]
    \centering
    \begin{tabular}{|c|c|c|}
    \hline
    \textbf{Optical Flow} & \textbf{Median Filtering} & \textbf{Kernelized Correlation Filters} \\ \hline
    9.8 FPS               & 4.7 FPS                   & 2.1 FPS                                 \\ \hline
    \end{tabular}
    \vspace{2mm}
    \caption{Comparing average detector-tracker processing speeds over the dataset.}
    \label{table:trackerspeeds} 
    \vspace{-3mm}
\end{table}

\subsection{Generating heat maps of frequently visited places}

Figure \ref{fig:heatmaps} illustrates the resulting heat maps that indicate the most frequently visited spots in the store. This gives a clear indication to the shop owners which counters attract customers better than others. Generating heat maps with the detector unit only tends to generate heat maps with visually only very small differences. While the tracker helps in obtaining higher processing speeds in capturing some of the missed detections due to occlusion, it does not really benefit the heatmap generation part.

\vspace{-1mm}
\section{Conclusion} \label{conclusion}
\vspace{-1mm}

In this work we propose a solution for mapping the flow of customers in shop-like environments in real-time based on person detection and tracking, implemented on an power-efficient and compact embedded platform. In this
work we proposed a novel batch processing approach that efficiently uses the power of both GPU and CPU simultaneously, to run a state-of-the-art TensorRT optimized YOLOv3 object detector combined with a sparse optical flow object tracker on the embedded Jetson TX2 platform. We achieve an average precision of 81.59\% at a processing speed of 10 FPS on a dataset of challenging real-life imagery acquired in a real shop. While this is not yet the optimal solution, given the challenging conditions (dynamic backgrounds, illumination changes and different store lay-outs), this numbers are quite impressive. On top of that we provide visually pleasing heat maps of the store, giving owners valuable insights to customer behavior. To improve the current results, we are planning to add a person re-identification pipeline to the solution and keep an eye out to new ways of optimizing deep learning pipelines for embedded platforms. This paper creates a proof-of-concept setup that can be further exploited in multiple camera stream setups, exploring how we can use shared memory buffers and shared detection networks between different embedded platforms efficiently.

\vspace{-1mm}
\section{Acknowledgements} \label{acknowledgements}
\vspace{-1mm}

This work is supported by PixelVision, KU Leuven and Flanders Innovation \& Entrepreneurship (VLAIO) through a Baekelandt scholarship.

\bibliographystyle{splncs04}
{\small
\bibliography{acivs2020.bib}}

\end{document}